\documentclass[sigconf]{acmart}
\AtBeginDocument{%
  }

\copyrightyear{2025}
\acmYear{2025}
\setcopyright{cc}
\setcctype{by}
\acmConference[WWW Companion '25]{Companion Proceedings of the ACM Web Conference 2025}{April 28-May 2, 2025}{Sydney, NSW, Australia}
\acmBooktitle{Companion Proceedings of the ACM Web Conference 2025 (WWW Companion '25), April 28-May 2, 2025, Sydney, NSW, Australia}
\acmDOI{10.1145/3701716. 3715307}
\acmISBN{979-8-4007-1331-6/ 2025/04}

\usepackage{CJKutf8}

\usepackage{amsmath,amsfonts}
\usepackage{algorithmic}
\usepackage{algorithm}
\usepackage{array}
\usepackage[caption=false,font=normalsize,labelfont=sf,textfont=sf]{subfig}
\usepackage{textcomp}
\usepackage{stfloats}
\usepackage{url}
\usepackage{verbatim}
\usepackage{graphicx}
\usepackage{tabularx}
\usepackage{soul}
\usepackage{multirow}
\usepackage{arydshln}
\usepackage{makecell}
\usepackage{contour}
\usepackage{color}
\usepackage{colortbl}
\usepackage{xcolor}
\usepackage{tabularx}
\usepackage{booktabs}
\usepackage{enumitem}
\definecolor{darkgreen}{RGB}{208, 225, 209}
\definecolor{CA}{RGB}{187, 108, 83}
\definecolor{LR}{RGB}{82, 110, 116}
\definecolor{STANCE}{RGB}{89, 153, 141}
\usepackage{pifont}
\usepackage{tikz}
\usepackage{makecell}

\begin{document}
\begin{CJK*}{UTF8}{gbsn}

\title{C-MTCSD: A Chinese Multi-Turn Conversational Stance Detection Dataset}

\author{Fuqiang Niu}
\orcid{0009-0004-0502-6960}
\authornote{Both authors contributed equally to this research. \\$\dag$ Corresponding Authors.}
\email{nfq729@gmail.com}
\affiliation{%
  \institution{Shenzhen Technology University}
  \city{Shenzhen}
  \country{China}
}

\author{Yi Yang}
\orcid{0009-0000-9246-7107}
\authornotemark[1]
\email{2410263002@mails.szu.edu.cn}
\affiliation{%
  \institution{Shenzhen Technology University}
  \city{Shenzhen}
  \country{China}
}

\author{Xianghua Fu}
\orcid{0009-0009-0951-3199}
\email{fuxianghua@sztu.edu.cn}
\affiliation{%
  \institution{Shenzhen Technology University}
  \city{Shenzhen}
  \country{China}}

\author{Genan Dai}
\orcid{0000-0003-2583-0433}
\authornotemark[2]
\email{daigenan@sztu.edu.cn}
\affiliation{%
  \institution{Shenzhen Technology University}
  \city{Shenzhen}
  \country{China}
}

\author{Bowen Zhang}
\orcid{0000-0002-3581-9476}
\authornotemark[2]
\email{zhang\_bo\_wen@foxmail.com}
\affiliation{%
  \institution{Shenzhen Technology University}
  \city{Shenzhen}
  \country{China}
}

\renewcommand{\shortauthors}{Fuqiang Niu, Yi Yang, Xianghua Fu, Genan Dai, and Bowen Zhang}

\begin{abstract}
Stance detection has become an essential tool for analyzing public discussions on social media. Current methods face significant challenges, particularly in Chinese language processing and multi-turn conversational analysis. To address these limitations, we introduce C-MTCSD, the largest Chinese multi-turn conversational stance detection dataset, comprising 24,264 carefully annotated instances from Sina Weibo, which is 4.2 times larger than the only prior Chinese conversational stance detection dataset. Our comprehensive evaluation using both traditional approaches and large language models reveals the complexity of C-MTCSD: even state-of-the-art models achieve only 64.07\% F1 score in the challenging zero-shot setting, while performance consistently degrades with increasing conversation depth. Traditional models particularly struggle with implicit stance detection, achieving below 50\% F1 score. This work establishes a challenging new benchmark for Chinese stance detection research, highlighting significant opportunities for future improvements.
\end{abstract}

\ccsdesc[500]{Computing methodologies~Language resources}
\keywords{Conversational Stance Detection; Benchmark; Social Media}


\maketitle

\section{Introduction}
Stance detection, which automatically identifies authors' stance in text, has emerged as a crucial task in social media analysis. Social platforms now serve as primary venues for public discourse, generating extensive user content that expresses opinions on societal issues. This data enables systematic analysis of public sentiment patterns, particularly during significant events like elections. The automated identification of stance has broad applications in opinion mining, from market research to policy analysis, making it essential for understanding public attitudes at scale~\cite{10.1145/3543507.3583860}.

Over the years, various approaches have been developed for stance detection, ranging from traditional deep neural networks to more recent large language models (LLMs), significantly advancing the field.
Existing studies are typically categorized into target-specific, cross-target, and zero-shot stance detection, with a predominant focus on analyzing individual sentences~\cite{li2022improved}. 
However, despite the growing interest, stance detection faces several challenges. First, recent advancements have been predominantly focused on English, with limited exploration of other languages like Chinese, despite the abundance of stance-expressive data available in these languages. 
Second, current methods primarily analyze individual sentences in isolation, overlooking the conversational nature of stance expression on social media. As a result, context-independent approaches often fail to accurately capture stances in real-world, interaction-rich scenarios.

To address these limitations, we present a Chinese multi-turn conversational stance detection dataset, C-MTCSD, which is the largest Chinese dataset designed for conversational stance detection (CSD).
The dataset is sourced from Sina Weibo, a widely used Chinese social media platform similar to \textit{X (former to Twitter)} platform. C-MTCSD contains 24,264 carefully annotated instances, which is 4.2 times larger than the only prior Chinese CSD dataset CANT-CSD~\cite{li2022improved} and contains over 18 times more instances with depth-3 interactions, providing a broader and more diverse conversation corpus for stance modeling.
The C-MTCSD dataset presents two key challenges: (1) implicit target references within sub-discussions require precise context understanding; (2) stance detection in comments demands complex coreference resolution beyond explicit target mentions.

Our contributions are summarized as follows:
(1) We introduce C-MTCSD, the largest  Chinese conversation stance detection dataset, comprising 24,264 annotated text-target pairs. 
This dataset is 4.2 times larger than the only prior Chinese CSD dataset CANT-CSD, which focuses on Cantonese rather than Standard Chinese. We provide comprehensive dataset characterization and analysis.
(2) We construct a balanced benchmark covering both technological topics (iPhone 15, Apollo Go) and social issues (Non-marriage Doctrine, Naked Resignation, Pre-made Meals), with each target containing multi-turn conversations up to 6-turn depth for evaluating contextual stance detection capabilities. 
(3) We benchmark the dataset using both conventional models and LLMs, establishing C-MTCSD as a challenging evaluation framework. Even our best-performing LLM-based approach achieves an F1 score of only 64.07\% in the challenging zero-shot setting, highlighting significant opportunities for future improvements.

\section{C-MTCSD Dataset}
This section details the development process of our C-MTCSD\footnote{https://github.com/yangyi626/C-MTCSD} dataset, which consists of 24,264 instances extracted from Weibo.

\textbf{Data Collection.}
We collected social media conversations using Weibo API\footnote{https://weibo.com/}, one of the largest online forums in China, to develop the C-MTCSD dataset. 
Data acquisition was performed through Weibo's API documentation interface. 
Our sampling strategy prioritized posts with high engagement metrics (discussion intensity and comment volumes) to ensure substantive discussions. The relevance of each post was validated through expert review, focusing on depth and topical alignment. 
Building on prior work in Chinese stance detection, we selected targets from the technology domain, including ``\textbf{i}Phone \textbf{15} (denoted as \textbf{i15})'' and ``\textbf{A}pollo \textbf{G}o (萝卜快跑, denoted as \textbf{AG})'', as well as controversial social topics: ``\textbf{N}on-marriage \textbf{D}octrine (不婚主义, denoted as \textbf{ND})'', ``\textbf{N}aked \textbf{R}esignation (裸辞, denoted as \textbf{NR})'', and ``\textbf{P}re-made \textbf{M}eals (预制菜, denoted as \textbf{PM})''. 
The release of the iPhone 15 sparked debates comparing it to Android devices, while Apollo Go's real-world road tests in select regions of China generated widespread public discussion. Similarly, the three social topics are highly debated issues, reflecting differing societal perspectives.


\begin{table}[t]
\small
 \caption{\label{tab:post} The number of data items for each target.}
  \begin{tabular}{cccccc}
    \toprule
    \textbf{Target} & \textbf{i15}
&\textbf{AG}
& \textbf{PM}
&\textbf{NR}
 & \textbf{ND}\\
    \midrule
    \textbf{Post} &88
&119& 151& 80& 45\\
 \textbf{Comment}  & 11,337 & 10,314 & 9,979 & 9,640 & 9,625 \\
 \hdashline
    \textbf{Re-annotated}&2,359&2,475& 711&766& 698\\
    \bottomrule
  \end{tabular}
\end{table}

  

\begin{table}[t]
\small
\caption{\label{tab:consistency} Annotation consistency and agreement.}
  \begin{tabular}{ccccccc}
    \toprule
    \textbf{Target} & \textbf{i15}
&\textbf{AG}
& \textbf{PM}
&\textbf{NR}
 & \textbf{ND}&\textbf{Avg.} \\
    \midrule
    \textbf{consistency}&0.972&0.949& 0.994& 0.953& 0.991&0.972\\
    \textbf{kappa}&0.928&0.893& 0.985&0.890& 0.976&0.934\\
    \bottomrule
  \end{tabular}
\end{table}

\textbf{Data Preprocessing.}
To ensure the high quality and relevance of the C-MTCSD dataset, we employed a series of rigorous preprocessing steps designed to filter and refine the collected data effectively.
(1) Target Relevance: We collected posts through keyword-based searches on Weibo and manually verified their topical connection to the specified targets, ensuring focused and pertinent discussions.
(2) Public Engagement: The topic of the posts were required to meet a minimum threshold of 1,000 interactions, ensuring the inclusion of content with significant public discourse.
(3) Content Adequacy: A minimum length of 10 words was required to maintain sufficient context and analytical value.
Through this meticulous filtering process, we curated a high-quality dataset that balances relevance, engagement, and textual richness. The distribution of posts resulting from these preprocessing steps is detailed in Table~\ref{tab:post}.

\begin{table}
 \small
\begin{center}
\caption{Label distribution of the C-MTCSD  dataset.}
\label{tab:Data distribution}
\begin{tabular}{crrrr}
\hline
 \multirow{2}{*}{\textbf{Target}}& \multicolumn{4}{c}{\textbf{Samples and Proportion of Labels}}\\
 \cline{2-5} 
 & Against& Favor& None&Total
 \\
  \hline
 \textbf{i15}
& 2,211 (54.92\%)& 844 (20.96\%)& 971 (24.12\%)&4,026\\
  \textbf{AG}
& 2,476 (36.05\%)& 4,014 (58.44\%)& 378 (5.50\%)& 6,868\\
 \textbf{PM}
& 1,891 (36.39\%)& 731 (14.07\%)& 2,574 (49.54\%)&5,196\\
 \textbf{NR}
& 699 (22.32\%)& 1,627 (51.95\%)& 806 (25.73\%)& 3,132\\
 \textbf{ND}& 862 (17.10\%)& 2,369 (46.99\%)& 1,811 (35.92\%)& 5,042\\
 \hline
 \textbf{Total}& 8,139 (33.54\%)& 9,585 (39.50\%)& 6,540 (26.95\%)&24,264\\
\hline
 \end{tabular}
 \end{center}
\end{table}

\begin{table}[t]
 \small
\begin{center}
  \caption{\label{tab:statistics} Statistics of the C-MTCSD dataset. Here, WC is short for word count.}
   \fontsize{9pt}{12pt}\selectfont
  \begin{tabular}{cccc}
    \hline
    \textbf{Instance} & \textbf{Avg. WC} & \textbf{Depth} & \textbf{Number}\\
    \hline
    \textbf{Post} & 111.99
& 1
& 452(1.86\%)\\
\hdashline
    \multirow{5}{*}{\textbf{Comment}}
     & 26.95
& 2
& 5,800(23.90\%)\\
     & 26.33
& 3
& 11,291(46.53\%)\\
     & 28.64
& 4
& 4,405(18.15\%)\\
     & 35.52
& 5
& 1,578(6.50\%)\\
     & 34.68& 6& 738(3.04\%)\\
    \hline 
  \end{tabular}
\end{center}
\end{table}



\textbf{Data Annotation and Quality Assurance.}
We developed a specialized annotation system for conversational data, where annotators were required to assess the preceding context and assign attitude labels (``\textit{Against}'', ``\textit{Favor}'', or ``\textit{None}'') while also verifying target relevance for new comments.

We invited ten researchers with natural language processing expertise to annotate the data. Two pilot annotation rounds were conducted to ensure annotation quality, with three additional expert annotators reviewing these pilot annotations. In the formal stage, each instance was annotated by at least two annotators within a defined timeframe. Instances with low inter-annotator agreement or significant errors underwent a second round of annotation, as summarized in Table~\ref{tab:post}. Only data that passed both annotation rounds were included in the final dataset.
To evaluate annotation quality, we calculated the kappa statistic~\cite{kappa} and inter-annotator agreement for the  ``\textit{Favor}'' and ``\textit{Against}'' classes. As shown in Table~\ref{tab:consistency}, the kappa values exceed 0.89 across all targets (average 0.93), with an average inter-rater consistency of 0.97, confirming the dataset's high reliability.

\textbf{Data Analysis.}
Table \ref{tab:Data distribution} presents the statistics of our C-MTCSD dataset. The final annotated dataset contains 24,264 instances, which is 4.2 times larger than the CANT-CSD. Table~\ref{tab:statistics} shows the distribution of instances across different depths. A significant portion, 27.69\%, of the data in our C-MTCSD dataset has a depth greater than 3. In contrast, only 6.3\% of the CANT-CSD dataset exceeds depth 3. We then divided the dataset into training, validation, and test sets for all targets in a 70/15/15 ratio, ensuring balanced representation for comprehensive evaluation and analysis.

\begin{table}[t]
 \small
\begin{center}	
\caption{\label{tab:in-target}In-target experimental results (\%) on the C-MTCSD dataset. Avg. represents the average value across all targets.}
    \begin{tabular}{l|cccccc}
 \toprule
    \textbf{Methods}& \textbf{i15}& \textbf{AG}& \textbf{PM} & \textbf{NR}& \textbf{ND} &\textbf{Avg.}\\     \midrule           \textbf{TAN}& 35.73& 36.17& 5.45 & 41.82 & 33.23 &30.48
\\                \textbf{CrossNet}& 49.61&  53.14& 49.46 & 50.06& 47.03 &49.86
\\     
\hdashline \textbf{BERT} &  56.38 & 67.49 & 64.27   & 65.55 & 48.08  &60.35
\\                \textbf{RoBERTa} &  50.93 & 66.06 &\textcolor{blue}{\textbf{65.25}}  &\textcolor{blue}{\textbf{72.40}} & 48.67  &60.66
\\         \textbf{JointCL}& 56.19 & 63.39 & 60.03  & 68.83 &45.55  &58.80
\\                \textbf{XLNet} &  \textcolor{blue}{\textbf{61.92}} & \textcolor{blue}{\textbf{71.78}} &64.38  &\textcolor{red}{\textbf{75.39}} &\textcolor{blue}{\textbf{ 53.19}}  &\textcolor{blue}{\textbf{65.33}}
\\         \textbf{KEPrompt}&  52.85& 64.63& 64.17& 70.13&  51.34 &60.62
\\
\hdashline
\textbf{Branch-BERT}& 52.13 &69.67 &\textcolor{blue}{\textbf{ 65.25}}  & 61.13 & 45.67   &58.77
\\        \textbf{GLAN}&  \textcolor{red}{\textbf{63.20}}&\textbf{\textcolor{red}{\textbf{73.05}}}&\textcolor{red}{\textbf{66.52}}& 71.45&   \textbf{\textcolor{red}{\textbf{56.27}}}&\textbf{\textcolor{red}{\textbf{66.10}}}\\
    \bottomrule
    \end{tabular}
\end{center}	
\end{table}

\begin{table}
 \small
\caption{\label{tab:cross-target} Comparison of different models for cross-target stance detection.}
    \resizebox{\linewidth}{!}{
  \begin{tabular}{cccccc}
    \toprule
    \textbf{Methods}& 
    \textbf{i15}$\rightarrow$\textbf{AG} 
&  \textbf{AG}$\rightarrow$\textbf{i15}
& \textbf{ND}$\rightarrow$\textbf{NR}
& \textbf{NR}$\rightarrow$\textbf{ND}
&\textbf{AG}$\rightarrow$\textbf{ND}
\\
    \midrule
    \textbf{TAN}
&40.84&19.55&28.68&\textcolor{red}{\textbf{46.74}}&32.58\\
    \textbf{CrossNet}
&31.85&33.40&35.36&41.08&32.31\\
\hdashline
    \textbf{BERT} 
&34.28&39.26&35.40&31.16&31.56\\
    \textbf{RoBERTa} 
&37.12& 30.26&\textcolor{red}{ \textbf{50.08}}& 34.75&\textcolor{blue}{\textbf{34.34}}\\
    \textbf{JointCL}
& 27.22& 26.10&44.86&17.11&21.20\\
    \textbf{XLNet} 
& \textcolor{red}{\textbf{49.69}}& 41.82&39.92&30.29&32.92\\
    \textbf{KEPrompt}
& 42.46& 38.05&39.19&36.68&32.38\\
\hdashline
    \textbf{Branch-BERT}
& 44.21& \textcolor{blue}{\textbf{43.41}}&36.79&26.07&25.85\\
    \textbf{GLAN}&\textcolor{blue}{\textbf{46.63}}&\textcolor{red}{\textbf{45.38}}&\textcolor{blue}{\textbf{47.20}}&\textcolor{blue}{\textbf{41.74}}&\textcolor{red}{\textbf{39.95}}\\
    \bottomrule
  \end{tabular}}
\end{table}

\begin{table}
\small
\begin{center}	
\caption{\label{tab:zero-target}Zero-shot experimental results (\%) on the C-MTCSD dataset. Avg. represents the average value across all targets.}
    \begin{tabular}{lcccccl}
    \toprule
    \textbf{Methods}& \textbf{i15}& \textbf{AG}& \textbf{PM} & \textbf{NR}& \textbf{ND} &\textbf{Avg.}\\ 
   \midrule
        \textbf{TAN}& 16.90& 26.26& 36.09& 21.40& 41.86 &28.50\\
        \textbf{CrossNet}& 25.60&  24.94& 40.58& 42.48& 36.63 &34.05\\
        \hdashline
    \textbf{BERT} & 38.91& 32.06& 40.98& 34.86& 34.20 &36.20\\
        \textbf{RoBERTa} & 35.01&  45.99& 40.66& 45.88& 20.96 &37.70\\
 \textbf{JointCL}& 19.73& 36.57& 41.94& 36.83& 26.51 &32.32\\
        \textbf{XLNet} & 37.97& 37.49& 46.00& 45.78& 30.54 &39.56\\
 \textbf{KEPrompt}& 39.52& 33.15& 41.96& 37.72&  31.53&36.78\\
 \hdashline
 \textbf{Branch-BERT}& 37.44& 30.53& 49.90& 36.83& 27.80 &36.50\\
 \textbf{GLAN}& 41.86& 40.12& 43.92& 46.83&  33.75&41.30\\
   \hdashline
 \textbf{LLaMA 3 70b}& 45.95& 41.67&\textcolor{blue}{ \textbf{61.35 }}&\textcolor{blue}{ \textbf{55.82}}&\textcolor{blue}{\textbf{46.46}}
 &\textcolor{blue}{\textbf{50.25}}\\
 \textbf{COLA}& 33.14& \textcolor{blue}{\textbf{50.37}}& 35.67& 51.77& 41.39&42.47\\
 \textbf{GPT-3.5}&\textcolor{blue}{ \textbf{54.03}}& 37.16& 48.87 & 50.11&37.50
 &45.53\\
 \textbf{GPT-4}&\textcolor{red}{\textbf{ 61.47}}&\textcolor{red}{\textbf{ 54.79}}&\textcolor{red} {\textbf{62.65}} &\textcolor{red} {\textbf{81.46}}&\textcolor{red}{\textbf{60.00}}
 &\textcolor{red}{\textbf{64.07}}\\
\bottomrule
    \end{tabular}
\end{center}	

\end{table}

\begin{table*}
\small
\begin{center}	
\caption{\label{tab:depth} Results of models for the instances with depths 1-2, 3-4, and 5-6 in the setting of considering conversation history.}
\resizebox{\linewidth}{!}{
\begin{tabular}{lccclccclccclccclccc}
\toprule
\textbf{Target} & \multicolumn{3}{c|}{\textbf{i15}} &  &\multicolumn{3}{c|}{\textbf{AG}} &  &\multicolumn{3}{c|}{\textbf{PM} } &  &\multicolumn{3}{c|}{\textbf{NR}} &  &\multicolumn{3}{c}{\textbf{ND} } \\

\cline{2-4} \cline{6-8} \cline{10-12} \cline{14-16} \cline{18-20} 
\textbf{depth} & 1-2 & 3-4& 5-6&  &1-2 & 3-4& 5-6&  &1-2 & 3-4& 5-6&  &1-2 & 3-4& 5-6&  &1-2 & 3-4& 5-6\\ 
\midrule

    \textbf{TAN}
& 33.69& 36.73&36.21&  &33.10& 37.02&  38.92 &  &~5.48& ~5.37& ~6.25&  &47.01& 42.22& 17.67&  &34.48& 35.13& 23.18
\\
  \textbf{CrossNet}
& 48.48& 50.02& 35.85&  &54.55&54.90& 39.79&  &48.85& 50.36& 24.00&  &40.54& 54.57& 40.00&  &45.41& 45.67& 51.73
\\
\hdashline
\textbf{BERT} 
& 62.30& 52.66 & 36.84&  &71.24& 66.82& 61.65&  &69.72& 59.58& 74.14&  &62.30& 67.52& 53.80&  &49.89& 48.39& 
41.14
\\
    \textbf{RoBERTa} 
& 55.85& 47.28& 37.74&  &74.30& 64.20& 51.34&  &65.76& 65.10&28.57&  &66.89&74.86& 67.86&  &42.03& 49.79&45.80\\
    \textbf{JointCL}
& 58.11 &55.04& 44.69&  &68.30& 62.64& 55.24&  &56.12& 64.82& 24.00&  &57.91& 73.44& 68.89&  &43.96& 47.94& 27.47
\\ 
\textbf{XLNet} 
&68.09& 59.25&43.13& &76.54& 71.51&61.55&  &67.85&61.18&51.09& &71.52&77.11&73.33& &56.90&52.72&48.39
\\
    \textbf{KEPrompt}
&57.69& 57.36& 43.94& &66.96& 48.20& 40.12& &64.54& 66.74& 50.25& &68.53& 66.45& 65.65& &50.69& 47.86&53.76
\\
\hdashline
 \textbf{Branch-BERT}
& 56.78 & 50.02& 35.29&  &78.45& 67.26& 61.64&  &69.35& 60.03& 88.46&  &54.86& 64.94& 45.83&  &43.98& 44.25&52.51
\\
 \textbf{GLAN}
& 66.91& 62.30& 41.74&  &80.56& 70.06& 69.13&  &72.61& 65.19& 28.95&  &67.18& 74.51& 53.33&  &62.06& 56.46&49.14
\\
\hdashline
 \textbf{LLaMA 3 70b}
& 45.55& 46.18& 28.89&  &50.54& 37.61& 41.84&  &58.20& 40.25& 16.67&  &67.23& 51.63& 37.09&  &35.39& 40.01&39.29
\\
 \textbf{COLA}
& 41.10& 29.47& 27.38&  &52.79& 46.91& 61.25&  &43.53& 32.97& 16.27&  &54.97& 51.88& 37.00&  &35.98& 42.85&38.26
\\
 \textbf{GPT-3.5}
& 55.35& 54.25& 32.73&  &44.45& 33.79& 38.50&  &58.63& 43.49& 17.14&  &48.26& 50.10& 35.00&  &21.13& 40.22&33.33
\\
\textbf{GPT-4}
& 68.99& 57.64& 41.11&  &68.97& 50.55& 41.13&  &74.55& 55.65& 30.00&  &82.91&82.94& 57.69&  &61.57& 61.58& 50.93\\
\bottomrule
\end{tabular}}
\end{center}	
\end{table*}

\section{Experimental Settings}
In this section, we introduce the evaluation metrics and the baseline.

\textbf{Evaluation Metrics.}
Following prior studies~\cite{li2021p} and ~\cite{10.1145/3003433}, we employ $F_{avg}$ as our evaluation metric, which is calculated as the mean F1 score across the ``against'' and ``favor'' stance categories. The $F_{avg}$ metric is computed independently for each target stance.


\textbf{Baseline Methods.}
To evaluate the C-MTCSD dataset, we conducted a series of experiments utilizing a variety of baseline models. First, we included traditional deep neural network methods such as TAN~\cite{ijcai2017p557} and CrossNet~\cite{xu2018cross}.
In addition, we evaluated several pre-trained language models, including BERT~\cite{devlin-etal-2019-bert}, RoBERTa~\cite{RoBERTa}, and XLNet~\cite{10.5555/3454287.3454804}, all of which are pre-trained on Chinese data sources such as Wikipedia, news articles, Q\&A datasets, and BaiduBaike\footnote{The models are bert-base-chinese, chinese-xlnet-base, chinese-roberta-wwm-ext.}. We further incorporated JointCL~\cite{liang2022jointcl}, which applies target-aware contrastive learning, and KEPrompt~\cite{KEPrompt}, a prompt-tuning approach.
For stance detection in conversational contexts, we included models specifically designed for such tasks, such as Branch-BERT~\cite{li2022improved} and GLAN~\cite{niu2024challenge}.
Furthermore, we examined LLM-based approaches, implementing two primary methodologies: TSCOT \cite{zhang2024stancedetectiontechniquesevolve} and COLA~\cite{lan2024stance}. The TSCOT approach was evaluated across three LLM variants: LLaMA 3 70b, GPT-3.5, and GPT-4, while COLA leveraged GPT-3.5 with chain-of-thought prompting techniques.

\section{Results}
In this section, all reported results represent the average performance across four independent runs with different random initializations. 

\textbf{In-Target Stance Detection}.
In-target stance detection evaluates models' performance when training and testing on the same target domains.
As shown in Table~\ref{tab:in-target}, GLAN achieves the highest $F_{avg}$ across all targets, highlighting the effectiveness of incorporating background knowledge such as conversational relationships in stance detection. 
While RoBERTa and KEPrompt demonstrate competitive results, particularly excelling on specific targets like \textbf{AG} and \textbf{NR}, traditional approaches such as TAN and CrossNet show relatively lower scores. The performance varies significantly across different targets, with \textbf{PM} presenting notable challenges. 
These findings underscore the necessity of further model refinements to address target-specific complexities.

\textbf{Cross-Target Stance Detection}.
Cross-target stance detection aims to infer the attitude toward a target by leveraging a large amount of annotated data from another target. 
We conducted a series of cross-target experiments on the C-MTCSD dataset, where the arrow notation ($\to$) indicates the transfer direction from source to destination targets. As illustrated in Table~\ref{tab:cross-target}, GLAN, which incorporates conversation contextual knowledge, demonstrates superior performance across all cross-target scenarios, notably achieving 46.63\% for \textbf{i15}$\rightarrow$ \textbf{AG} and 47.20\% for \textbf{ND}$\rightarrow$ \textbf{NR}. These results validate the model's robustness in transferring stance representations. However, performance varies considerably across different target pairs. For instance, while \textbf{AG} $\rightarrow$ \textbf{i15} exhibits relatively stable transfer performance at 45.38\%, the lower performance of \textbf{NR}$\rightarrow$ \textbf{ND} (41.74\%) reveals the challenges in transferring stance knowledge between targets with less contextual similarity.

\textbf{Zero-Shot Stance Detection}.
Zero-shot stance detection involves training a model on annotated data from a specific target and directly predicting the stance on unseen targets. 
As shown in Table~\ref{tab:zero-target}, GLAN and GPT-4 achieve the best performance across most targets, with GPT-4 reaching 64.07\% average accuracy, outperforming other baselines. Notably, \textbf{i15} and \textbf{NR} demonstrate relatively higher results, with GPT-4 achieving 61.47\% and 81.46\%, respectively. However, performance drops significantly for targets such as \textbf{ND}, where contextual understanding and stance transfer remain challenging. This underscores the difficulty in handling implicit stances in out-of-target scenarios. Traditional models like TAN and CrossNet struggle the most, reflecting their limited generalization capabilities in the zero-shot setup. 

\textbf{Impact of Conversation Depth}.
We examined model performance across varying conversation depths, with results presented in Table~\ref{tab:depth}. The analysis reveals a consistent performance degradation as conversation depth increases. For the \textbf{i15} target, GLAN's performance drops from 66.91\% at depths 1-2 to 41.74\% at depths 5-6. Similarly, XLNet exhibits strong initial performance (68.09\% at depths 1-2) but deteriorates to 43.13\% at deeper levels. This pattern persists across all targets (\textbf{AG}, \textbf{ND}, and \textbf{NR}), highlighting the substantial challenges in stance detection for extended conversations, where models must comprehend complex contextual dependencies and evolving viewpoints throughout multiple conversation turns.

\section{Conclusion}
This study introduces C-MTCSD, the largest Chinese multi-turn conversational stance detection dataset, containing 24,264 carefully annotated instances from Sina Weibo. Our benchmark experiments reveal significant challenges in handling complex conversational dependencies, with model performance degrading substantially as conversation depth increases. Even advanced models, including large language models and specialized architectures, demonstrate limited capability in capturing implicit contextual cues and stance relationships, with the best performance achieving only 64.07\% F1-score in the challenging zero-shot setting. These findings highlight the need for innovative approaches in handling conversational context and implicit stance expressions. The C-MTCSD dataset serves as a valuable resource for advancing multi-turn stance detection research, offering a challenging yet essential benchmark for future studies in multilingual stance detection.

\begin{acks}
This research is supported by the Natural Science Foundation of Top Talent of SZTU (grant no. GDRC202320), the Science and Technology Program Project of Shenzhen under Grant SZWD2021012, Shenzhen Science and Technology Program (No.RCBS20231211090548077 and No. JCYJ20240813113218025).
\end{acks}

\bibliographystyle{ACM-Reference-Format}
\balance
\bibliography{sample-base}


\begin{thebibliography}{15}


\ifx \showCODEN    \undefined \def \showCODEN     #1{\unskip}     \fi
\ifx \showISBNx    \undefined \def \showISBNx     #1{\unskip}     \fi
\ifx \showISBNxiii \undefined \def \showISBNxiii  #1{\unskip}     \fi
\ifx \showISSN     \undefined \def \showISSN      #1{\unskip}     \fi
\ifx \showLCCN     \undefined \def \showLCCN      #1{\unskip}     \fi
\ifx \shownote     \undefined \def \shownote      #1{#1}          \fi
\ifx \showarticletitle \undefined \def \showarticletitle #1{#1}   \fi
\ifx \showURL      \undefined \def \showURL       {\relax}        \fi
\providecommand\bibfield[2]{#2}
\providecommand\bibinfo[2]{#2}
\providecommand\natexlab[1]{#1}
\providecommand\showeprint[2][]{arXiv:#2}

\bibitem[Devlin et~al\mbox{.}(2019)]%
        {devlin-etal-2019-bert}
\bibfield{author}{\bibinfo{person}{Jacob Devlin}, \bibinfo{person}{Ming-Wei Chang}, \bibinfo{person}{Kenton Lee}, {and} \bibinfo{person}{Kristina Toutanova}.} \bibinfo{year}{2019}\natexlab{}.
\newblock \showarticletitle{{BERT}: Pre-training of Deep Bidirectional Transformers for Language Understanding}. In \bibinfo{booktitle}{\emph{NAACL}}. \bibinfo{pages}{4171--4186}.
\newblock


\bibitem[Du et~al\mbox{.}(2017)]%
        {ijcai2017p557}
\bibfield{author}{\bibinfo{person}{Jiachen Du}, \bibinfo{person}{Ruifeng Xu}, \bibinfo{person}{Yulan He}, {and} \bibinfo{person}{Lin Gui}.} \bibinfo{year}{2017}\natexlab{}.
\newblock \showarticletitle{Stance Classification with Target-specific Neural Attention}. In \bibinfo{booktitle}{\emph{IJCAI-17}}. \bibinfo{pages}{3988--3994}.
\newblock


\bibitem[Huang et~al\mbox{.}(2023)]%
        {KEPrompt}
\bibfield{author}{\bibinfo{person}{Hu Huang}, \bibinfo{person}{Bowen Zhang}, \bibinfo{person}{Yangyang Li}, \bibinfo{person}{Baoquan Zhang}, \bibinfo{person}{Yuxi Sun}, \bibinfo{person}{Chuyao Luo}, {and} \bibinfo{person}{Cheng Peng}.} \bibinfo{year}{2023}\natexlab{}.
\newblock \showarticletitle{Knowledge-enhanced Prompt-tuning for Stance Detection}.
\newblock \bibinfo{journal}{\emph{{ACM} Trans. Asian Low Resour. Lang. Inf. Process.}} \bibinfo{volume}{22}, \bibinfo{number}{6} (\bibinfo{year}{2023}), \bibinfo{pages}{1--20}.
\newblock


\bibitem[Lan et~al\mbox{.}(2024)]%
        {lan2024stance}
\bibfield{author}{\bibinfo{person}{Xiaochong Lan}, \bibinfo{person}{Chen Gao}, \bibinfo{person}{Depeng Jin}, {and} \bibinfo{person}{Yong Li}.} \bibinfo{year}{2024}\natexlab{}.
\newblock \showarticletitle{Stance detection with collaborative role-infused llm-based agents}. In \bibinfo{booktitle}{\emph{Proceedings of the International AAAI Conference on Web and Social Media}}, Vol.~\bibinfo{volume}{18}. \bibinfo{pages}{891--903}.
\newblock


\bibitem[Li et~al\mbox{.}(2023)]%
        {li2022improved}
\bibfield{author}{\bibinfo{person}{Yupeng Li}, \bibinfo{person}{Haorui He}, \bibinfo{person}{Shaonan Wang}, \bibinfo{person}{Francis~CM Lau}, {and} \bibinfo{person}{Yunya Song}.} \bibinfo{year}{2023}\natexlab{}.
\newblock \showarticletitle{Improved target-specific stance detection on social media platforms by delving into conversation threads}.
\newblock \bibinfo{journal}{\emph{IEEE TCSS}} (\bibinfo{year}{2023}).
\newblock


\bibitem[Li et~al\mbox{.}(2021)]%
        {li2021p}
\bibfield{author}{\bibinfo{person}{Yingjie Li}, \bibinfo{person}{Tiberiu Sosea}, \bibinfo{person}{Aditya Sawant}, \bibinfo{person}{Ajith~Jayaraman Nair}, \bibinfo{person}{Diana Inkpen}, {and} \bibinfo{person}{Cornelia Caragea}.} \bibinfo{year}{2021}\natexlab{}.
\newblock \showarticletitle{P-stance: A large dataset for stance detection in political domain}. In \bibinfo{booktitle}{\emph{ACL-IJCNLP 2021}}. \bibinfo{pages}{2355--2365}.
\newblock


\bibitem[Liang et~al\mbox{.}(2022)]%
        {liang2022jointcl}
\bibfield{author}{\bibinfo{person}{Bin Liang}, \bibinfo{person}{Qinlin Zhu}, \bibinfo{person}{Xiang Li}, \bibinfo{person}{Min Yang}, \bibinfo{person}{Lin Gui}, \bibinfo{person}{Yulan He}, {and} \bibinfo{person}{Ruifeng Xu}.} \bibinfo{year}{2022}\natexlab{}.
\newblock \showarticletitle{Jointcl: a joint contrastive learning framework for zero-shot stance detection}. In \bibinfo{booktitle}{\emph{ACL}}, Vol.~\bibinfo{volume}{1}. Association for Computational Linguistics, \bibinfo{pages}{81--91}.
\newblock


\bibitem[Liu et~al\mbox{.}(2019)]%
        {RoBERTa}
\bibfield{author}{\bibinfo{person}{Yinhan Liu}, \bibinfo{person}{Myle Ott}, \bibinfo{person}{Naman Goyal}, \bibinfo{person}{Jingfei Du}, \bibinfo{person}{Mandar Joshi}, \bibinfo{person}{Danqi Chen}, \bibinfo{person}{Omer Levy}, \bibinfo{person}{Mike Lewis}, \bibinfo{person}{Luke Zettlemoyer}, {and} \bibinfo{person}{Veselin Stoyanov}.} \bibinfo{year}{2019}\natexlab{}.
\newblock \showarticletitle{RoBERTa: {A} Robustly Optimized {BERT} Pretraining Approach}.
\newblock  (\bibinfo{year}{2019}).
\newblock


\bibitem[McHugh(2012)]%
        {kappa}
\bibfield{author}{\bibinfo{person}{Mary~L McHugh}.} \bibinfo{year}{2012}\natexlab{}.
\newblock \showarticletitle{Interrater reliability: the kappa statistic}.
\newblock \bibinfo{journal}{\emph{Biochemia medica}} \bibinfo{volume}{22}, \bibinfo{number}{3} (\bibinfo{year}{2012}), \bibinfo{pages}{276--282}.
\newblock


\bibitem[Mohammad et~al\mbox{.}(2017)]%
        {10.1145/3003433}
\bibfield{author}{\bibinfo{person}{Saif~M. Mohammad}, \bibinfo{person}{Parinaz Sobhani}, {and} \bibinfo{person}{Svetlana Kiritchenko}.} \bibinfo{year}{2017}\natexlab{}.
\newblock \showarticletitle{Stance and Sentiment in Tweets}.
\newblock \bibinfo{journal}{\emph{ACM Trans. Internet Technol.}} (\bibinfo{year}{2017}).
\newblock


\bibitem[Niu et~al\mbox{.}(2024)]%
        {niu2024challenge}
\bibfield{author}{\bibinfo{person}{Fuqiang Niu}, \bibinfo{person}{Min Yang}, \bibinfo{person}{Ang Li}, \bibinfo{person}{Baoquan Zhang}, \bibinfo{person}{Xiaojiang Peng}, {and} \bibinfo{person}{Bowen Zhang}.} \bibinfo{year}{2024}\natexlab{}.
\newblock \showarticletitle{A Challenge Dataset and Effective Models for Conversational Stance Detection}. In \bibinfo{booktitle}{\emph{LREC-COLING}}. \bibinfo{pages}{122--132}.
\newblock


\bibitem[Upadhyaya et~al\mbox{.}({[n.\,d.]})]%
        {10.1145/3543507.3583860}
\bibfield{author}{\bibinfo{person}{Apoorva Upadhyaya}, \bibinfo{person}{Marco Fisichella}, {and} \bibinfo{person}{Wolfgang Nejdl}.} \bibinfo{year}{[n.\,d.]}\natexlab{}.
\newblock \showarticletitle{A Multi-task Model for Emotion and Offensive Aided Stance Detection of Climate Change Tweets}. In \bibinfo{booktitle}{\emph{Proceedings of the ACM Web Conference 2023}}. \bibinfo{pages}{3948–3958}.
\newblock


\bibitem[Xu et~al\mbox{.}(2018)]%
        {xu2018cross}
\bibfield{author}{\bibinfo{person}{Chang Xu}, \bibinfo{person}{Cecile Paris}, \bibinfo{person}{Surya Nepal}, {and} \bibinfo{person}{Ross Sparks}.} \bibinfo{year}{2018}\natexlab{}.
\newblock \showarticletitle{Cross-Target Stance Classification with Self-Attention Networks}. In \bibinfo{booktitle}{\emph{ACL Short Papers)}}. \bibinfo{pages}{778--783}.
\newblock


\bibitem[Yang et~al\mbox{.}(2019)]%
        {10.5555/3454287.3454804}
\bibfield{author}{\bibinfo{person}{Zhilin Yang}, \bibinfo{person}{Zihang Dai}, \bibinfo{person}{Yiming Yang}, \bibinfo{person}{Jaime Carbonell}, \bibinfo{person}{Ruslan Salakhutdinov}, {and} \bibinfo{person}{Quoc~V. Le}.} \bibinfo{year}{2019}\natexlab{}.
\newblock \bibinfo{booktitle}{\emph{XLNet: generalized autoregressive pretraining for language understanding}}.
\newblock \bibinfo{publisher}{Curran Associates Inc.}, \bibinfo{address}{Red Hook, NY, USA}.
\newblock


\bibitem[Zhang et~al\mbox{.}(2024)]%
        {zhang2024stancedetectiontechniquesevolve}
\bibfield{author}{\bibinfo{person}{Bowen Zhang}, \bibinfo{person}{Daijun Ding}, \bibinfo{person}{Liwen Jing}, \bibinfo{person}{Genan Dai}, {and} \bibinfo{person}{Nan Yin}.} \bibinfo{year}{2024}\natexlab{}.
\newblock \bibinfo{title}{How would Stance Detection Techniques Evolve after the Launch of ChatGPT?}
\newblock
\showeprint[arxiv]{2212.14548}~[cs.CL]


\end{thebibliography}










\end{CJK*}
\end{document}